\documentclass{article}

\usepackage[arxiv,final]{neurips_2026}

\usepackage[dvipsnames]{xcolor}
\usepackage{microtype}
\usepackage{hyperref}
\usepackage{url}
\usepackage{booktabs}
\usepackage{todonotes}
\usepackage{booktabs} 
\usepackage{graphicx}
\usepackage{subcaption}
\usepackage{amssymb} 
\usepackage{enumitem}
\usepackage{amsmath}
\usepackage[table]{xcolor}
\usepackage{bbm}
\usepackage{fancyhdr}
\usepackage{multirow}

\newcommand{\system}{\textsc{Pace}}
\newcommand{\benchmark}{\textsc{Pace-Bench}}
\newcommand{\numsources}{19}
\newcommand{\numabilities}{11}
\newcommand{\nummodels}{14}
\newcommand{\numagentic}{4}
\newcommand{\numsamples}{100}
\usepackage{xcolor}

\usepackage{lineno}
\usepackage{color-edits}
\addauthor{gn}{magenta}

\definecolor{darkblue}{rgb}{0, 0, 0.5}
\hypersetup{colorlinks=true, citecolor=darkblue, linkcolor=darkblue, urlcolor=darkblue}

\title{\system{}: A Proxy for Agentic Capability Evaluation}

\author{%
  Yueqi Song$^{1}$, Lintang Sutawika$^{1}$, Jiarui Liu$^{1}$, Lindia Tjuatja$^{1}$, Jiayi Geng$^{1}$, Yunze Xiao$^{1}$, \\
  \textbf{Daniel Lee$^{2}$, Aditya Bharat Soni$^{1}$, Vincent Lo$^{1}$, Xiang Yue$^{1}$, Graham Neubig$^{1}$} \\[1.0ex]
  $^{1}$Carnegie Mellon University \quad $^{2}$Salesforce AI Research \\[0.5ex]
  \texttt{\{yueqis, gneubig\}@cs.cmu.edu} \\[1.0ex]
  \href{https://github.com/neulab/pace}{\includegraphics[height=0.4cm]{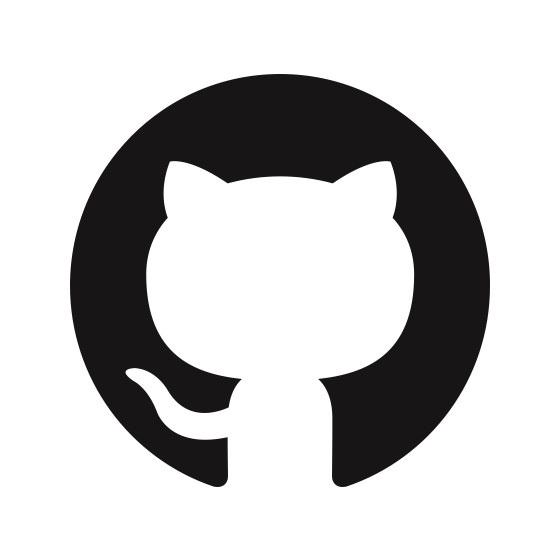}~\textbf{neulab/pace}}
  \quad
  \href{https://huggingface.co/datasets/neulab/pace-bench}{\includegraphics[height=0.4cm]{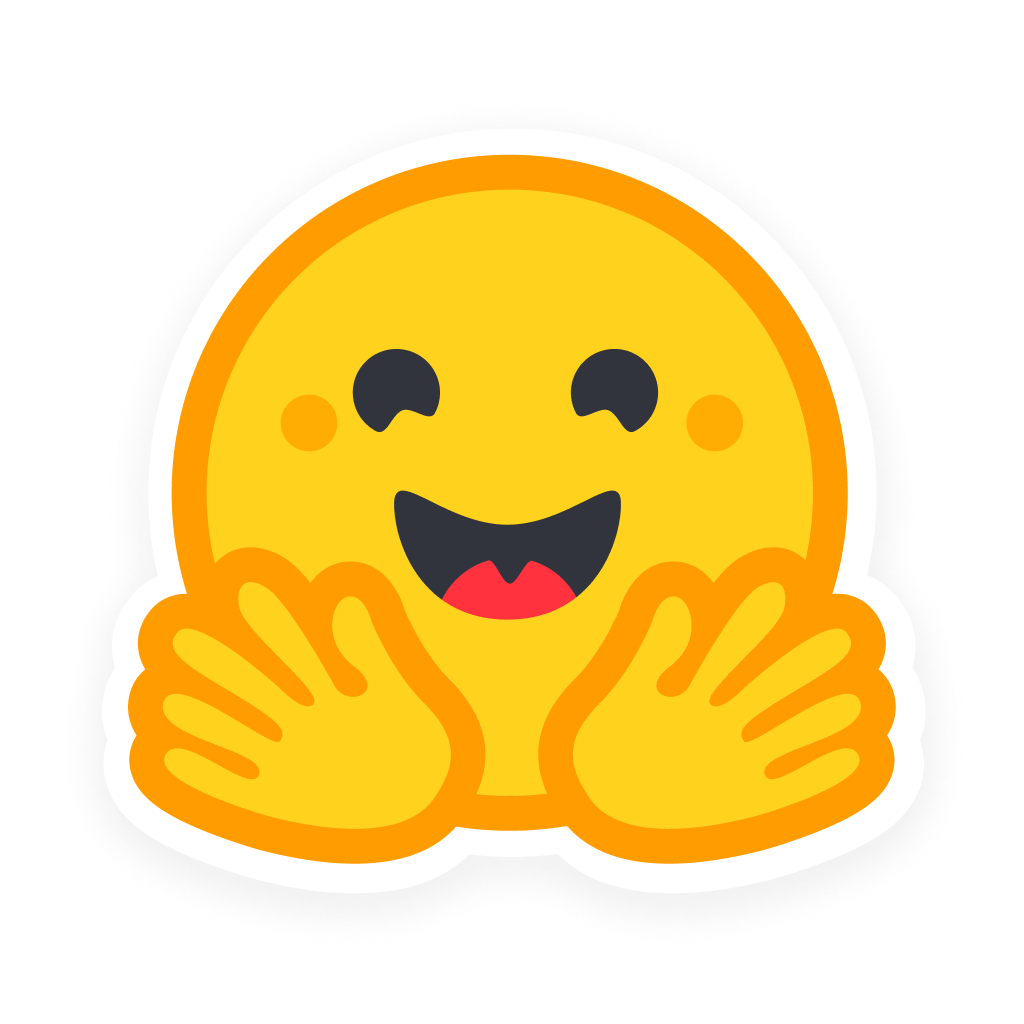}~\textbf{neulab/pace-bench}}
}

\begin{document}

\maketitle
\thispagestyle{fancy}
\fancyhead{}
\lhead{\includegraphics[height=0.5cm]{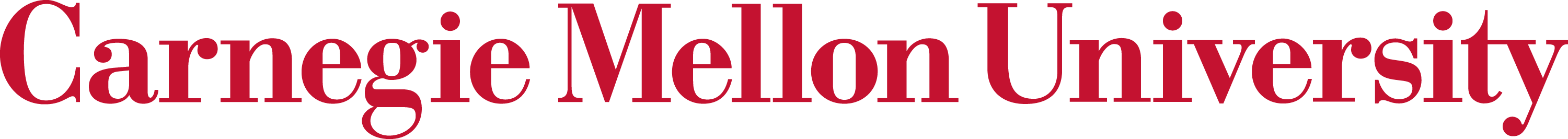}}
\rhead{\raisebox{-0.1cm}{\includegraphics[height=0.8cm]{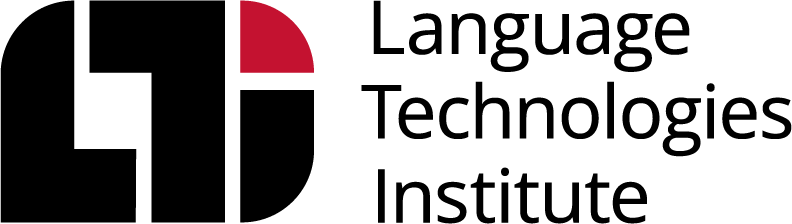}}}
\renewcommand{\headrulewidth}{0pt}
\setlength{\headheight}{12pt}
\setlength{\headsep}{3mm}

\vspace{-15pt}
\begin{abstract}

Evaluating large language model (LLM) agents on benchmarks like SWE-Bench and GAIA can be expensive, time-consuming, and requires complex infrastructure. A single evaluation can cost thousands of dollars and take days to complete. In contrast, non-agentic LLM benchmarks that test individual capabilities (e.g., reasoning, code generation, instruction following) are fast and cheap to run. In this paper, we investigate whether performance on expensive agentic benchmarks can be accurately predicted by the performance on a small, carefully selected subset of atomic evaluation instances.
We introduce \system{}, a framework that constructs proxy benchmarks by selecting instances from existing non-agentic evaluations whose aggregate scores most reliably predict model performances on agentic benchmarks. Given a pool of candidate instances spanning atomic capabilities (instruction following, planning, tool calling, etc.), \system{} fits a regression that maps a model's scores on a compact subset of source instances to its score on the target agentic benchmark. The subset itself is curated by combining two complementary instance-selection strategies, target-relevance local selection and globally informative global selection. 
We apply \system{} to the \numagentic{} target agentic benchmarks in this paper, which yields \benchmark{}, the concrete proxy benchmark that we evaluate in the paper.
Experiments across \nummodels{} models, \numagentic{} agentic benchmarks, and \numsources{} non-agentic benchmarks show that \benchmark{} predicts agentic scores with leave-one-out cross-validation (LOOCV) mean absolute error (MAE) under $4\%$, Spearman correlation above $0.80$, and pairwise model-ranking accuracy around $85\%$, all at much less than $1\%$ of the full agentic evaluation cost. We further analyze the selected proxy instances, revealing which skills each agentic benchmark uniquely demands. \system{} enables practitioners to obtain reliable estimates of agentic performance during model development, selection, and routing, without the overhead of full agent evaluation.

\end{abstract}

\begin{figure}[!h]
\centering
\vspace{-15pt}
\includegraphics[width=\textwidth]{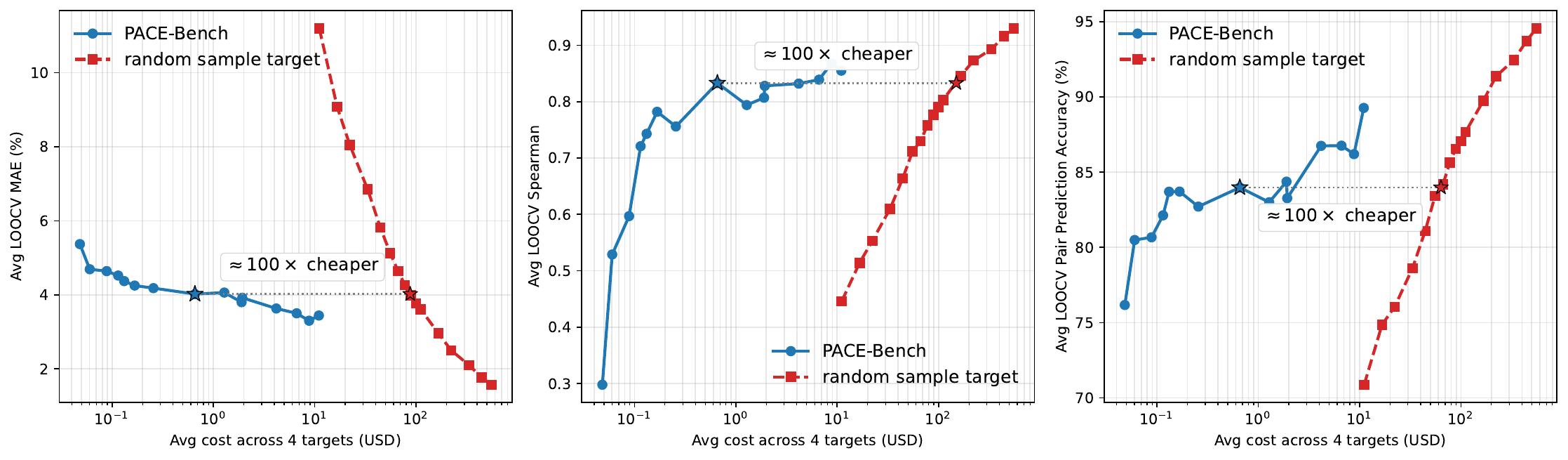}
\caption{\textbf{Cost-versus-quality tradeoff} of \system{} (blue) and sub-sampling target agentic evals (red), averaged across four datasets.
\textbf{Left:} mean absolute error. \textbf{Middle:} Spearman correlation. \textbf{Right:} pairwise model-ranking accuracy. At every budget below saturation, \system{} dominates sub-sampling agentic evals on all three metrics, matching quality at roughly $1/100$ of the cost.}
\label{fig:cost_curve}
\vspace{-20pt}
\end{figure}

\section{Introduction}

Tracking the progress of large language models (LLMs) capabilities has long relied on fast and inexpensive benchmarks that evaluate models' individual capabilities such as knowledge retrieval~\citep{hendryckstest2021, NEURIPS2024_ad236edc}, mathematical reasoning~\citep{hendrycksmath2021}, instruction following~\citep{zhou2023instructionfollowingevaluationlargelanguage}, code generation~\citep{jain2025livecodebench}, and more. 
Because such benchmarks consist of short, self-contained instances that can be scored with a single model invocation, these instances are cheap to run, easy to reproduce, and widely used for informing model development~\citep{liang2023holistic, biderman2024lessonstrenchesreproducibleevaluation}. 

As language models are increasingly deployed as agents, however, this evaluation paradigm breaks down. Agentic benchmarks such as SWE-Bench~\citep{jimenez2024swebench}, GAIA~\citep{mialon2024gaia}, and WebArena~\citep{zhou2024webarena} require models to operate over long horizons, interact with tools or environments, and recover from errors~\citep{Wang_2024}, often requiring complex infrastructure, long rollout times, and substantial API costs. Even evaluation of a single model under a single agent harness could cost thousands of dollars and take hours or days of setup and execution. These burdens often force researchers to evaluate models less frequently, report results on limited subsets, and make rigorous agent evaluation disproportionately accessible only to well-resourced groups. 

Despite the complexity of agentic tasks, success of LLMs on agentic benchmarks depends on model abilities like instruction following, planning, tool use, and reasoning, which are already measured by fast and inexpensive non-agentic benchmarks \citep{sumers2023cognitive,xi2025rise}. 
However, researchers still run full agent evaluations to compare models to track progress, suggesting that the predictive connection between model performance on non-agentic and agentic tasks is not yet well understood. Thus, we ask: \emph{can non-agentic benchmarks serve as a reliable and low-cost proxy for agentic benchmarks}?

To answer this question, we propose \system{} (\underline{P}roxy for \underline{A}gentic \underline{C}apability \underline{E}valuation), a simple yet effective framework that selects a compact subset of non-agentic benchmark instances whose aggregate scores could best predict the target agentic benchmark performances across models.
\system{} draws its candidate pool of non-agentic evaluation instances from existing benchmarks that broadly cover skills that intuitively seem important for agentic tasks, such as instruction following, tool calling, multimodal understanding, etc.
We formulate the construction of \system{} as a budget-constrained subset selection problem. Given a fixed budget of $C$ proxy instances, \system{} uses a calibration set of models with known scores on both the candidate pool and a target agentic benchmark to identify which $C$ instances are the most predictive of the target benchmark.
Concretely, \system{} fits a least-squares regression that maps each model's per-instance scores on the $C$ selected source instances to its target benchmark mean, with the calibration models supplying training data; bootstrap resampling over the target instances stabilizes the regression weights against label noise.
The $C$ instances themselves are produced by combining two complementary criteria, a target-relevance local signal (rank-correlation with target labels) and a globally-informative global signal (SVD leverage in the source matrix).

Our approach differs from prior benchmark compression and subset-selection methods~\citep{perlitz-etal-2024-efficient, 10.5555/3692070.3693466}, which aim to reduce costs within a single target benchmark, and from approaches that recast agent tasks into alternative formats (e.g., multiple choice questions)~\citep{qin2025aptbenchbenchmarkingagenticpotential}. Instead, we seek to predict model performances on a target agentic benchmark using a compact subset drawn from a separate candidate pool of inexpensive evaluation instances, with no modification to how the target benchmark is scored.

To demonstrate the empirical effectiveness of \system{}, we evaluate \system{} across \nummodels{} models, \numagentic{} agentic benchmarks (GAIA \citep{mialon2024gaia}, SWE-Bench Multimodal \citep{yang2025swebench}, SWE-Bench Verified \citep{jimenez2024swebench}, SWT-Bench \citep{mundler2024swtbench}), and \numsources{} source non-agentic benchmarks spanning \numabilities{} capabilities of LLMs.

\autoref{fig:cost_curve} summarizes our answer: a small, well-chosen subset of non-agentic instances tracks agentic performance closely, at less than $\frac{1}{100}$ of the cost of either a full agent evaluation or a random subset of the target benchmark itself. Concretely, using a proxy of just \numsamples{} instances, \system{} achieves strong predictive performance.  At equal prediction quality, \system{} requires roughly $100\times$ less cost in dollars than a random target-sampling baseline. Our main findings are as follows: 
\begin{itemize}[leftmargin=*]

    \item \textbf{Generalization to Unseen Models:} Instances selected with \system{} on a training set of models generalize to a held-out set of models not seen during selection. This setting is relevant to deployment, where the goal is to predict a new model's agentic performance, without committing to full agentic evaluation. Specifically, across the \numagentic{} benchmarks, \system{} predicts agentic scores with leave-one-out cross-validation (LOOCV) mean absolute error (MAE) under $4\%$, Spearman correlation above $0.80$, and pairwise model-ranking accuracy around $85\%$, all much less than $1\%$ of the full agentic evaluation cost.
    \item \textbf{Predictable Cost-Accuracy Tradeoff:} We show that the cost-accuracy tradeoff is smooth and highly predictable. As shown in \autoref{fig:cost_curve}, prediction quality broadly improves with the proxy budget and then saturates, with diminishing returns past a few hundred instances; even small budgets are already highly competitive. This lets practitioners choose evaluation budgets that match their specific resource constraints.
    \item \textbf{Interpretability of Capabilities:} By examining the number of selected instances from each benchmark and each model capability, we identify which model capabilities most strongly affect each target agentic task, providing interpretable evidence for the capability structure that underlies successes and failures in agentic tasks. 
\end{itemize}

\section{Background}

\begin{table}[t]
\centering
\resizebox{\textwidth}{!}{%
\setlength{\tabcolsep}{2.5pt}
\begin{tabular}{l *{11}{c} c r l}
\toprule
\textbf{Benchmark} & \textbf{IF} & \textbf{LCA} & \textbf{ER} & \textbf{Plan} & \textbf{Code} & \textbf{IR} & \textbf{CS} & \textbf{TC} & \textbf{Reas} & \textbf{MM} & \textbf{Ver} & \textbf{Setup} & \textbf{\#Inst} & \textbf{Cost} \\
\midrule
\multicolumn{15}{c}{\textbf{Agentic Benchmarks}} \\
\midrule
\textbf{GAIA}                   & $\bullet$ & $\bullet$ & $\bullet$ & $\bullet$ & $\bullet$ & $\bullet$  &           & $\bullet$ & $\bullet$ & $\bullet$  & $\bullet$ & \textcolor{red}{\textbf{H}} & 165     & \$ 0.38 \\
\textbf{SWE-Bench Multimodal}   & $\bullet$ & $\bullet$ & $\bullet$ & $\bullet$ & $\bullet$ &            & $\bullet$ & $\bullet$ & $\bullet$ & $\bullet$  & $\bullet$ & \textcolor{red}{\textbf{H}} & 102     & \$ 1.89 \\
\rowcolor{gray!15}
\textbf{SWE-Bench Verified}     & $\bullet$ & $\bullet$ & $\bullet$ & $\bullet$ & $\bullet$ &            & $\bullet$ & $\bullet$ & $\bullet$ &            & $\bullet$ & \textcolor{red}{\textbf{H}} & 500   & \$ 1.19 \\
\textbf{SWT-Bench}              & $\bullet$ & $\bullet$ & $\bullet$ & $\bullet$ & $\bullet$ &            &  $\bullet$   &           & $\bullet$ &            & $\bullet$ & \textcolor{red}{\textbf{H}} & 430   & \$ 0.98 \\
\midrule
\multicolumn{15}{c}{\textbf{Non-Agentic Benchmarks}} \\
\midrule
\textbf{ACPBench}               & $\bullet$ &           &           & $\bullet$ &           &            &           &           & $\bullet$ &            & $\bullet$ & \textcolor{OliveGreen}{\textbf{L}} & 1,040      & \$ 0.009 \\
\rowcolor{gray!15}
\textbf{AIME 2025}              & $\bullet$ &           &           &           &           &            &           &           & $\bullet$ &            &           & \textcolor{OliveGreen}{\textbf{L}}  & 30      & \$ 0.015 \\
\textbf{BEIR (NFCorpus)}        & $\bullet$ &           &           &           &           & $\bullet$  &           &           &  &            &           & \textcolor{OliveGreen}{\textbf{L}}          & 323      & \$ 0.121 \\
\rowcolor{gray!15}
\textbf{BFCL}                   & $\bullet$ &           &           &  &           &            &           & $\bullet$ & $\bullet$ &            & $\bullet$ & \textcolor{OliveGreen}{\textbf{L}}      & 5,343   & \$ 0.008 \\
\textbf{DebugBench}             & $\bullet$ &           & $\bullet$ &           & $\bullet$ &            &           &           & $\bullet$ &            & $\bullet$ & \textcolor{YellowOrange}{\textbf{M}}   & 4,253   & \$ 0.005 \\
\rowcolor{gray!15}
\textbf{GPQA}                   & $\bullet$ &           &           &           &           &            &           &           & $\bullet$ &            &           & \textcolor{OliveGreen}{\textbf{L}}          & 2,384     & \$ 0.009 \\
\textbf{HumanEval}              & $\bullet$ &           &           &           & $\bullet$ &            &           &           & $\bullet$ &            & $\bullet$ & \textcolor{OliveGreen}{\textbf{L}}          & 164     & \$ 0.004 \\
\rowcolor{gray!15}
\textbf{IFEval}                 & $\bullet$ &           &           &           &           &            &           &           & $\bullet$ &            &           & \textcolor{OliveGreen}{\textbf{L}}          & 541     & \$ 0.006 \\
\textbf{InFoBench}              & $\bullet$ &           &           &           &           &            &           &           & $\bullet$ &            &           & \textcolor{OliveGreen}{\textbf{L}}          & 500   & \$ 0.009 \\
\rowcolor{gray!15}
\textbf{LIFBench}               & $\bullet$ & $\bullet$ &           &           &           &            &           &           & $\bullet$ &            &           & \textcolor{OliveGreen}{\textbf{L}}          & 2,766   & \$ 0.118 \\
\textbf{LiveCodeBench}          & $\bullet$ &           & $\bullet$ & $\bullet$ & $\bullet$ &            &           &           & $\bullet$ &            & $\bullet$ & \textcolor{YellowOrange}{\textbf{M}}    & 2,870     & \$ 0.051 \\
\rowcolor{gray!15}
\textbf{LogiQA}                 & $\bullet$ &           &           &           &           &            &           &           & $\bullet$ &            &           & \textcolor{OliveGreen}{\textbf{L}}          & 1,302   & \$ 0.007 \\
\textbf{MBPP}                   & $\bullet$ &           &           &           & $\bullet$ &            &           &           & $\bullet$ &            & $\bullet$ & \textcolor{OliveGreen}{\textbf{L}}          & 500     & \$ 0.004 \\
\rowcolor{gray!15}
\textbf{MMLU}                   & $\bullet$ &           &           &           &           &            &           &           & $\bullet$ &            &           & \textcolor{OliveGreen}{\textbf{L}}          & 28,084  & \$ 0.006 \\
\textbf{MMMU}                   & $\bullet$ &  &           &           &           &            &           &           & $\bullet$ & $\bullet$  &           & \textcolor{OliveGreen}{\textbf{L}}          & 900  & \$ 0.012 \\
\rowcolor{gray!15}
\textbf{PlanBench}              & $\bullet$ &           &           & $\bullet$ &           &            &           &           & $\bullet$ &            & $\bullet$ & \textcolor{OliveGreen}{\textbf{L}}          & 4,000      & \$ 0.007 \\
\textbf{RepoBench-R}              & $\bullet$ & $\bullet$ &           &           & $\bullet$ &            & $\bullet$ &           & $\bullet$ &            &           & \textcolor{YellowOrange}{\textbf{M}}    & 2,010 & \$ 0.013 \\
\rowcolor{gray!15}
\textbf{VisualPuzzles}          & $\bullet$ &  &           &           &           &            &           &           & $\bullet$ & $\bullet$  &           & \textcolor{OliveGreen}{\textbf{L}}          & 1,168   & \$ 0.018 \\
\textbf{VisualWebBench}         & $\bullet$ &  &           &           &           & $\bullet$  &           &           & $\bullet$ & $\bullet$  &           & \textcolor{OliveGreen}{\textbf{L}}          & 1,536   & \$ 0.007 \\

\bottomrule
\end{tabular}%
}
\caption{Overview of existing agentic and non-agentic benchmarks, with capability coverage, setup complexity, number of instances, and estimated per-instance evaluation cost for Claude Sonnet 4.5.
\textbf{IF}=Instruction Following, \textbf{LCA}=Long Context Aggregation, \textbf{ER}=Error Recovery, \textbf{Plan}=Planning, \textbf{Code}=Code Generation, \textbf{IR}=Information Retrieval, \textbf{CS}=Code Search, \textbf{TC}=Tool Calling, \textbf{Reas}=Reasoning, \textbf{MM}=Multimodal Understanding, \textbf{Ver}=Verification and Test. We classify a setup effort for each benchmark within 3 classifications of \textcolor{Red}{\textbf{High}} (requiring setting up evaluation environments), \textcolor{YellowOrange}{\textbf{Medium}} (medium effort setup) or \textcolor{OliveGreen}{\textbf{Low}} (only needing API calls).}
\label{tab:benchmarks_combined}
\vspace{-20pt}
\end{table}

\subsection{From Static to Agentic Benchmarks}
\label{subsec:capabilities}
Standard LLM evaluation has historically relied on atomic, single-turn, and largely static benchmarks that are inexpensive to run and easy to reproduce.
In contrast, evaluating LLM-based agents requires models to act over longer horizons, interact with tools or environments, and recover from intermediate errors. The agentic benchmarks often require sandboxes, browsers, repositories, or custom runtime environments, and can be sensitive to environmental noise, harness design, and external dependencies~\citep{fan2025sweeffireevaluatingsoftwareai}.
What makes prediction plausible across these two evaluation protocols is that they require and evaluate models on overlapping underlying capabilities. We organize these into \numabilities{} categories: instruction following, long context aggregation, error recovery, planning, code generation, information retrieval, code search, tool calling, reasoning, multimodal understanding, and verification and test\footnote{\autoref{app:benchmark_details} includes citations and descriptions of these benchmarks and capabilities.}. \autoref{tab:benchmarks_combined} maps each non-agentic and agentic benchmark to the capabilities it requires, alongside its evaluation cost and setup requirements. SWE-Bench \citep{jimenez2024swebench}, for instance, requires models to possess capabilities like planning, code generation, and verification, all of which are also measured by cheaper non-agentic benchmarks, suggesting its agentic performance is in principle predictable from non-agentic signals.

Recent work suggests that complex model-level behavior can be partially predicted from simpler evaluation signals. Collaborative Performance Prediction (CPP)~\citep{zhang-etal-2024-collaborative} uses matrix factorization to estimate missing model-task outcomes, ONEBench~\citep{ghosh-etal-2025-onebench} studies sample-level unification across open-ended capabilities, and several meta-analyses have shown that aggregated static benchmark scores can predict human-preference-based Elo~\citep{spangher-etal-2025-chatbot, ramaswamy-etal-2025-model}. Although human preference and agentic execution are substantively different outcomes, this literature supports the broader premise that model performance exhibits predictable structure across evaluations. This motivates the question: if agentic success depends on model capabilities already measured by cheaper non-agentic benchmarks, a carefully selected subset of those instances may serve as a reliable proxy.

\subsection{Problem Formulation}
\label{sec:formulation}

Let $M = \{m_1, \ldots, m_{|M|}\}$ be a set of $|M|$ language models and $T = \{t_1, \ldots, t_{|T|}\}$ be a target agentic benchmark with $|T|$ instances. For each model $m$ and index $i \in \{1, \ldots, |T|\}$, let $y_{m,i} \in [0, 1]$ denote $m$'s score on instance $t_i \in T$. Collecting over all models and instances gives the target score matrix $Y \in [0, 1]^{|M| \times |T|}$. We write $\bar y_m = \frac{1}{|T|} \sum_{i=1}^{|T|} y_{m,i}$ for model $m$'s mean score on the target benchmark.
Let also $\mathcal{S} = \{S_1, \ldots, S_{|\mathcal{S}|}\}$ be a set of $|\mathcal{S}|$ non-agentic source benchmarks, where each benchmark $S$ has $|S|$ instances.
For each model $m$, source benchmark $S$, and index $j \in \{1, \ldots, |S|\}$, let $X_S[m, j] \in [0, 1]$ denote $m$'s (possibly non-binary) score on the $j$-th instance of $S$. Collecting these gives the per-benchmark source score matrix $X_S \in [0,1]^{|M| \times |S|}$.

For a held-out evaluation protocol, we partition $M$ into a \emph{calibration} set $M_\mathrm{train}$ and an \emph{evaluation} set $M_\mathrm{eval}$, with calibration models used to fit the selection and predictor, and held-out models used to measure generalization.

\textbf{Goal A (Performance Prediction).}
Given a budget $C \in \mathbb{N}$ of source instances, select a set of instance indices $P = \bigsqcup_{S \in \mathcal{S}} P_S$ from the source pool, where $P_S \subseteq \{1, \ldots, |S|\}$ is the set of selected indices for source benchmark $S$, such that $|P| = \sum_{S \in \mathcal{S}} |P_S| = C$. Then learn a predictor $f$ s.t.
\begin{equation}
    \hat y_m = f\!\left(\left\{X_S[m, j]\right\}_{j \in P_S,\; S \in \mathcal{S}}\right) \approx \bar y_m,
    \label{eq:goal}
\end{equation}
where the input to $f$ is model $m$'s scores on every selected instance across all source benchmarks. This goal answers the question \emph{what score would a model achieve on the target agentic benchmark?}

\textbf{Goal B (Pairwise Preference Prediction).}
Using the same selection process, learn a predictor $g$ that, for any ordered pair $(m, m') \in M \times M$ with $m \neq m'$, outputs a binary label indicating which model is stronger on the target benchmark:

\begin{equation}
    \hat{b}_{m,m'} = g\!\left(\left\{X_S[m, j],\, X_S[m', j]\right\}_{j \in P_S,\; S \in \mathcal{S}}\right) \in \{0, 1\},
    \quad\text{with}\quad \hat{b}_{m,m'} \approx \mathbbm{1}[\bar y_m > \bar y_{m'}].
    \label{eq:goal_pair}
\end{equation}
This goal answers the question \emph{which of two models is stronger on the target agentic benchmark?}

\section{\system{}: A Proxy for Agentic Capability Evaluation}
\label{sec:method}

\begin{figure}[t]
\centering
\includegraphics[width=0.7\textwidth]{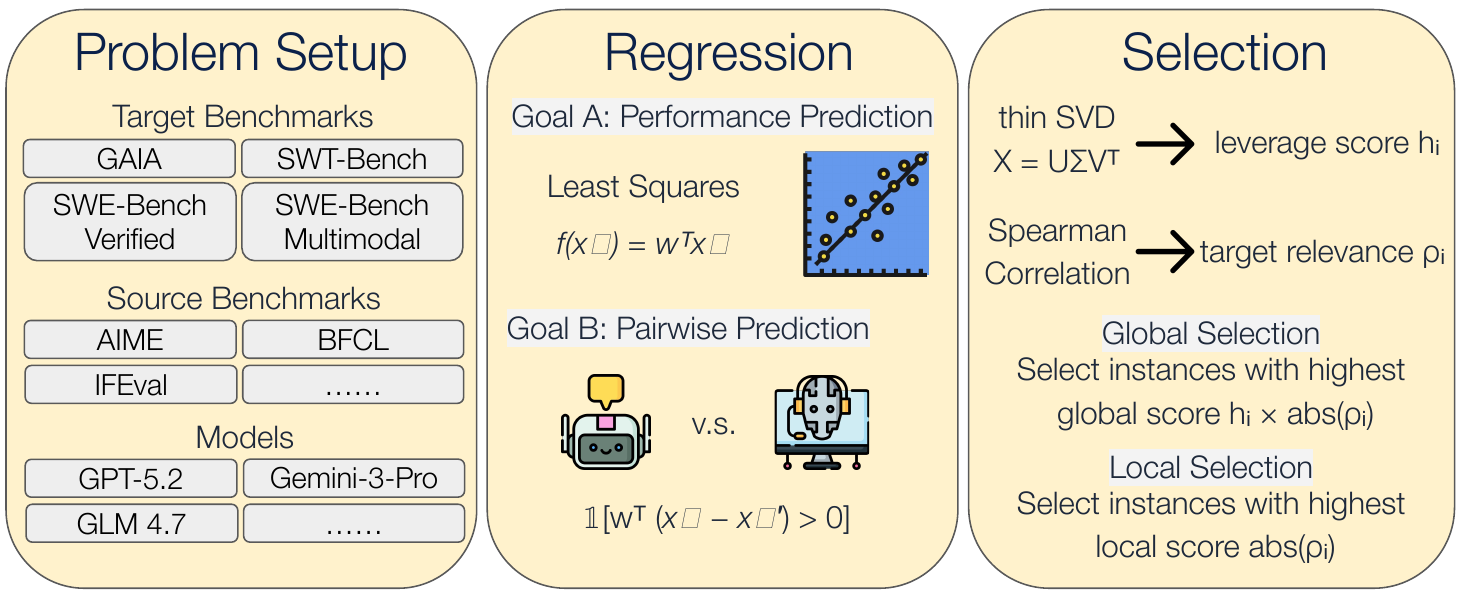}
\caption{Overview of \system{}. From a pool of non-agentic source benchmarks, two complementary filter-based criteria (Local: target relevance; Global: SVD leverage $\times$ relevance) each pick $C$ instances; the selected scores then drive a noise-aware regression that predicts the target agentic benchmark's mean score (Goal A) and pairwise model preferences (Goal B).}
\label{fig:method}
\end{figure}


Building on the problem formulation in \autoref{sec:formulation}, we describe \system{}, our framework for effectively and efficiently predicting model performances (Goal A) and pairwise model preferences (Goal B) on a target agentic benchmark using non-agentic benchmarks. \autoref{fig:method} is a high level overview of \system{}.

\system{} consists of two core components: (1) a \textbf{regression} that, given selected source instances for each target benchmark, builds a noise-aware predictor for either goals (described in \autoref{sec:regression}); and (2) an \textbf{instance selection} method that selects instances via two complementary strategies, \textbf{Local} and \textbf{Global} selection (\autoref{sec:selection}). We describe regression first, because the evaluation criterion for selection depends on how the selected instances are used downstream.

\subsection{Regression}
\label{sec:regression}

Our regression predicts a vector $x_m \in \mathbb{R}^{C}$ collecting model $m$'s scores on the $C$ source instances selected for the target (selection is described in \autoref{sec:selection}). The supervision signal is the target mean $\bar y_m = \frac{1}{|T|} \sum_{i=1}^{|T|} y_{m,i}$.

\paragraph{Goal A (Performance Prediction).}
We fit a linear least-squares regression $f(x_m) = w^\top x_m$ over the training models $m \in M_\mathrm{train}$, with the coefficient vector $w \in \mathbb{R}^{C}$ obtained by minimising $\sum_{m \in M_\mathrm{train}} \bigl(w^\top x_m - \bar y_m\bigr)^2$ (with regularization hyperparameters tuned by held-out evaluation). The prediction for a held-out model $m \in M_\mathrm{eval}$ is
\begin{equation}
    \hat y_{m} \,=\, w^\top x_{m}.
    \label{eq:goal-a-pred}
\end{equation}

\paragraph{Goal B (Pairwise Preference Prediction).}
For each ordered pair $(m, m') \in M_\mathrm{train} \times M_\mathrm{train}$ with $m \neq m'$, we form the source-score difference $x_m - x_{m'}$ and the label $\mathbbm{1}[\bar y_m > \bar y_{m'}]$. We fit a logistic regressor $g(\Delta x) = w_g^{\!\top} \Delta x$ (a structured Bradley-Terry model~\citep{firth2005bradley}) on these pair-differences; for a pair $(m, m')$ involving the held-out model $m$, the predicted logit is
\begin{equation}
    \hat z_{m\!,m'} \,=\, w_g^{\!\top} (x_{m} - x_{m'}),
    \label{eq:goal-b-pred}
\end{equation}
and the binary outcome is $\hat b_{m\!,m'} = \mathbbm{1}[\hat z_{m\!,m'} > 0]$.

\paragraph{Bootstrapping target instances.}
Agentic evaluation is expensive, so each target benchmark contains only a few hundred instances, along with a small model pool. Both scarcities make the target mean $\bar y_m$ a noisy estimate of model $m$'s true target performance, and treating it as exact lets the regression weights underestimate predictive uncertainty and overfit to one target-instance sample. We therefore draw $B$ bootstrap replicates of each $\bar y_m$ by resampling target instances with replacement, yielding $\bar y_m^{(b)}$ for $b = 1, \ldots, B$. Replacing $\bar y_m$ in the Goal A least-squares and Goal B logistic training objectives defined above with the concatenation of these replicates lets the regression average over the sampling distribution, making the weights estimate less sensitive to target-instance sampling noise.

\subsection{Instance Selection}
\label{sec:selection}

We now turn to selecting the proxy subset $P$ of size $C$ defined in \autoref{sec:formulation}. To enable cross-benchmark selection, we stack the per-benchmark score matrices into a single source matrix $X = [X_{S_1} \mid X_{S_2} \mid \cdots \mid X_{S_{|\mathcal{S}|}}] \in [0,1]^{|M| \times |\mathcal{I}|}$ indexed by $\mathcal{I} = \bigsqcup_{S \in \mathcal{S}} \{1, \ldots, |S|\}$, which enumerates all candidate instances across benchmarks (so $|\mathcal{I}| = \sum_{S \in \mathcal{S}} |S|$).

Given this setup, a natural approach is to fit a regularized linear model on $X$ and use the resulting Lasso \citep{tibshirani1996lasso} or Ridge \citep{hoerl1970ridge} weights to select the proxy instances. However, in our regime ($|M_\mathrm{train}| \ll |\mathcal{I}|$) the joint fit is severely under-determined and overfits (see \autoref{app:sparse_baseline} for more details). We thus decouple selection from regression.

\paragraph{Decoupling selection from regression via SVD.}
We instead score each source instance \emph{independently} using two complementary filter-based signals, both cheap to compute and stable under perturbations of $M_\mathrm{train}$. To define them, we perform a thin SVD on $X$: $X = U \Sigma V^\top$ ($U \in \mathbb{R}^{|M| \times n_c}$, $V \in \mathbb{R}^{|\mathcal{I}| \times n_c}$, where $n_c$ is the SVD rank hyperparameter). The two signals are:
\begin{itemize}[leftmargin=*]
\item \emph{Geometric importance}: the leverage score of instance $i \in \mathcal{I}$ in the SVD latent space, defined as $h_i \triangleq \sum_c V_{c,i}^2$, which measures its contribution to the global latent structure of the source pool, a classical criterion for selecting informative columns of a matrix~\citep{mahoney2009cur}.
\item \emph{Target relevance}: $\mathrm{abs}(\rho_i) \triangleq \mathrm{abs}\bigl(\mathrm{Spearman}(X_{M_\mathrm{train}, i},\, \bar y_{M_\mathrm{train}})\bigr)$, i.e., the rank consistency between the instance score and the target mean across training models. This is a standard filter-based feature-selection criterion \citep{guyon2003introduction}.
\end{itemize}

These two signals represent a standard tradeoff of \emph{shared geometric prior} versus \emph{task specialization} \citep{saeys2007review}, and \system{} draws on both. We \emph{jointly} select a single proxy subset $P = L \cup G$ with $|P| = C$, splitting the budget into two per-strategy sub-budgets $C_L + C_G = C$:

\begin{itemize}[leftmargin=*]
    \item \emph{Global} subset ($G$). This strategy jointly considers geometric importance and target relevance, and selects the top-$C_G$ instances by the product of the two signals: $\sigma_i \,=\, h_i \,\times\, \mathrm{abs}(\rho_i)$. Here $V$ is target-independent and serves as a \emph{prior} over the global latent structure of the source pool: leverage $h_i = \sum_c V_{c,i}^2$ identifies information-rich instances, with target relevance applied as a secondary filter. The held-out model $m^\star \in M_\mathrm{eval}$ is not in the SVD decomposition, so we obtain its latent-space coordinates separately. We project its score row $X_{m^\star, :}$ onto $V$ via the pseudoinverse $V^{+}$. This places $m^\star$ in the same $n_c$-dimensional latent space as the training models, allowing the regression to operate uniformly across all models.

    \item \emph{Local} subset ($L$). This strategy selects the top-$C_L$ instances solely by target relevance $\mathrm{abs}(\rho_i)$ (without using any geometric signal), and then \emph{recomputes} the SVD on the $|M| \times |L|$ submatrix $X_L$ to obtain a local basis $V_\mathrm{loc}$. As in Global selection, the held-out model $m^\star$ is projected into this basis via the pseudoinverse $V_\mathrm{loc}^{+}$, applied to its restricted score row $X_{m^\star, L}$ over the selected $C_L$ instances. The resulting embedding lives in the \emph{local} SVD space adapted to the selected subset rather than the global pool.
\end{itemize}

The two subsets are complementary: Global alone may select high-leverage instances that are irrelevant to the target, whereas Local alone ignores any geometric structure outside the selected subset. \system{} therefore uses both, combining them at prediction time rather than at selection time.

When the two top-ranked sets overlap, $|L \cup G|$ falls short of $C$; we greedily extend each side past columns already in $L \cup G$ (in proportion to the $C_L : C_G$ split) until $|L \cup G| = C$, so only $C$ unique source instances are evaluated per new model. \system{}'s final output is the ensemble $\hat y = \lambda \cdot \hat y_L + (1-\lambda) \cdot \hat y_G$, with both hyperparameters the ensemble weight $\lambda$ and the budget split $(C_L, C_G)$ optimized through held-out validation, letting the data rather than prior assumptions determine these ratios.

\section{Experiments and Results}
\label{sec:experiments}

\subsection{Experimental Setup}

\paragraph{Target Agentic Benchmarks.}
We evaluate \system{} on all \numagentic{} agentic benchmarks from \autoref{tab:benchmarks_combined}. These benchmarks vary in terms of tasks, required model abilities, and model performances; together they provide broad coverage of agentic tasks, including browser-based question answering, repository-level code generation, multimodal software engineering, and test understanding and construction.

\begin{itemize}[leftmargin=*]
\item \textbf{GAIA}~\citep{mialon2024gaia} is a benchmark for general-purpose AI assistants, consisting of real-world questions that require reasoning, tool use, web browsing, and, in some cases, multimodal understanding. It evaluates whether agents can decompose underspecified information-seeking tasks, gather evidence from external sources, and produce concise answers.

\item \textbf{SWE-Bench Verified}~\citep{jimenez2024swebench} evaluates repository-level software engineering agents on real GitHub issues. Given a codebase and an issue description, an agent must localize the relevant code, implement a patch, and pass the corresponding tests, making it a benchmark for practical bug fixing and code modification.

\item \textbf{SWE-Bench Multimodal}~\citep{yang2025swebench} extends software-engineering evaluation to issues that contain visual information, such as screenshots, mockups, diagrams, or visually presented error messages. It tests whether agents can combine visual understanding with repository-level code reasoning to resolve realistic software issues.

\item \textbf{SWT-Bench}~\citep{mundler2024swtbench} evaluates agents on software test generation for real-world bug fixes. Given the original codebase and a user-reported issue, the agent must generate tests that expose the bug by failing before the fix and passing after the fix, thereby measuring its ability to understand intended behavior and construct effective validation tests.

\end{itemize}

All agentic benchmark results are obtained using the OpenHands Index~\citep{openhandsindex2025}, which evaluates models with the OpenHands Software Agent SDK~\citep{wang2025openhandssoftwareagentsdk} under a unified agent harness, enabling fair cross-model comparisons.

\paragraph{Source Non-Agentic Benchmarks.}
The candidate source pool consists of all \numsources{} non-agentic benchmarks from \autoref{tab:benchmarks_combined}. These benchmarks collectively cover all \numabilities{} capabilities identified in \autoref{subsec:capabilities}. We evaluate non-agentic benchmarks using \texttt{lm-evaluation-harness}~\citep{eval-harness}; for benchmarks not yet supported by this package, we use each benchmark's official evaluation code.

\paragraph{Models.}
We evaluate across \nummodels{} models spanning proprietary and open models, including GPT 5.2 \citep{GPT5.2}, GPT 5.2 Codex \citep{GPT5.2Codex}, Gemini 3 Pro Preview \citep{gemini3}, Gemini 3 Flash Preview \citep{gemini3}, Claude Opus 4.5 \citep{Claude4.5Opus} and 4.6 \citep{Claude4.6Opus}, Claude Sonnet 4.5 \citep{Claude4.5Sonnet}, DeepSeek V3.2 \citep{liu2025deepseek}, GLM 4.7 \citep{GLM4.7}, Kimi K2~\citet{kimiteam2026kimik2openagentic} and K2.5~\citet{kimiteam2026kimik25visualagentic}, MiniMax M2.1 \citep{MiniMaxM2.1} and M2.5 \citep{MiniMaxM2.5}, and Qwen3 Coder 480B A35B \citep{Qwen3Coder}.

\paragraph{Evaluation Protocol.}
We evaluate \system{} under a strict \textbf{LOOCV (leave-one-out cross-validation)} protocol. In each fold, one of the \nummodels{} models is held out as $m^*$, while the remaining models are used for source-instance selection and regression. We then aggregate the held-out predictions across all \nummodels{} folds. This protocol answers the central question: \emph{can \system{} generalize to unseen models?}

\subsection{Main Results}
\label{sec:main_results}

Table~\ref{tab:main_results} reports the predictive power of \system{} with $C = 100$ proxy instances across all four agentic benchmarks for both Goal A and Goal B.

\begin{table}[htbp]
\centering

\begin{tabular}{l cccc}
\toprule
& \multicolumn{3}{c}{\textbf{Goal A}} & \textbf{Goal B} \\
\cmidrule(lr){2-4} \cmidrule(lr){5-5}
\textbf{Target} & \textbf{MAE} & \textbf{Spearman Corr.} & \textbf{Pearson Corr.} & \textbf{Accuracy} \\
\midrule
GAIA                 & 5.77\% & 0.79 & 0.78 & 83.33\% \\
SWE-bench Verified   & 2.09\% & 0.67 & 0.61 & 78.54\% \\
SWE-bench Multimodal & 2.23\% & 0.89 & 0.80 & 85.39\% \\
SWT-bench            & 5.12\% & 0.89 & 0.78 & 90.11\% \\
\midrule
\textbf{Average}     & \textbf{3.80\%} & \textbf{0.81} & \textbf{0.74} & \textbf{84.37\%} \\
\bottomrule
\end{tabular}

\caption{Strict LOOCV results on Goal A (absolute score prediction; MAE, Spearman, and Pearson correlation) and Goal B (pairwise preference prediction; accuracy).}
\label{tab:main_results}
\end{table}

\paragraph{Goal A (Performance Prediction).}\system{} achieves an average MAE of $3.80\%$ and a Spearman correlation of $0.807$ on Goal A, showing that only $100$ non-agentic source instances are sufficient to accurately predict absolute scores on agentic benchmarks. Although GAIA and SWT-bench exhibit slightly larger absolute errors, they still show strong ranking performance (Spearman $\geq 0.79$), suggesting that under LOOCV the dominant signal is model-capability ordering rather than precise numerical calibration.

\paragraph{Goal B (pairwise preference prediction).}
Reusing the same selected instances under a pairwise logistic, \system{} achieves an average LOOCV pair-accuracy of $84.4\%$, far above the random $50\%$ baseline.

\subsection{Trade Off of Performance v.s. Cost of \benchmark{}}
\label{sec:budget_sweep}
\autoref{fig:cost_curve} plots the cost-versus-quality tradeoff of \system{} (blue) against a random target-sampling baseline (red), averaged across the four agentic targets. The left plot reports LOOCV MAE (lower is better), the middle one reports LOOCV Spearman correlation (higher is better), and the right one reports pairwise model-ranking accuracy (higher is better). At every budget below saturation, \system{} dominates random target-sampling on all three metrics, matching \system{}'s quality with the random target-sampling baseline, but costing only roughly $\frac{1}{100}$ the cost of the baseline, demonstrating the effectiveness and efficiency of \system{}. We report the full budget sweep of \system{} as $C$ ranges from $25$ to $500$ in \autoref{app:budget_sweep}; quality broadly improves with the budget and then saturates, with $C=100$ sitting at a practical sweet spot.

\section{Analysis and Discussion}
\label{sec:analysis}

\subsection{What Does Learned Allocation Select?}
\label{sec:allocation_analysis}

\begin{figure}[htbp]
    \centering
    \caption{Total selected instances from source benchmarks covering each ability, for $C=100$ across the four agentic targets (abbreviations follow Table~\ref{tab:benchmarks_combined}).}
    \includegraphics[width=0.7\textwidth]{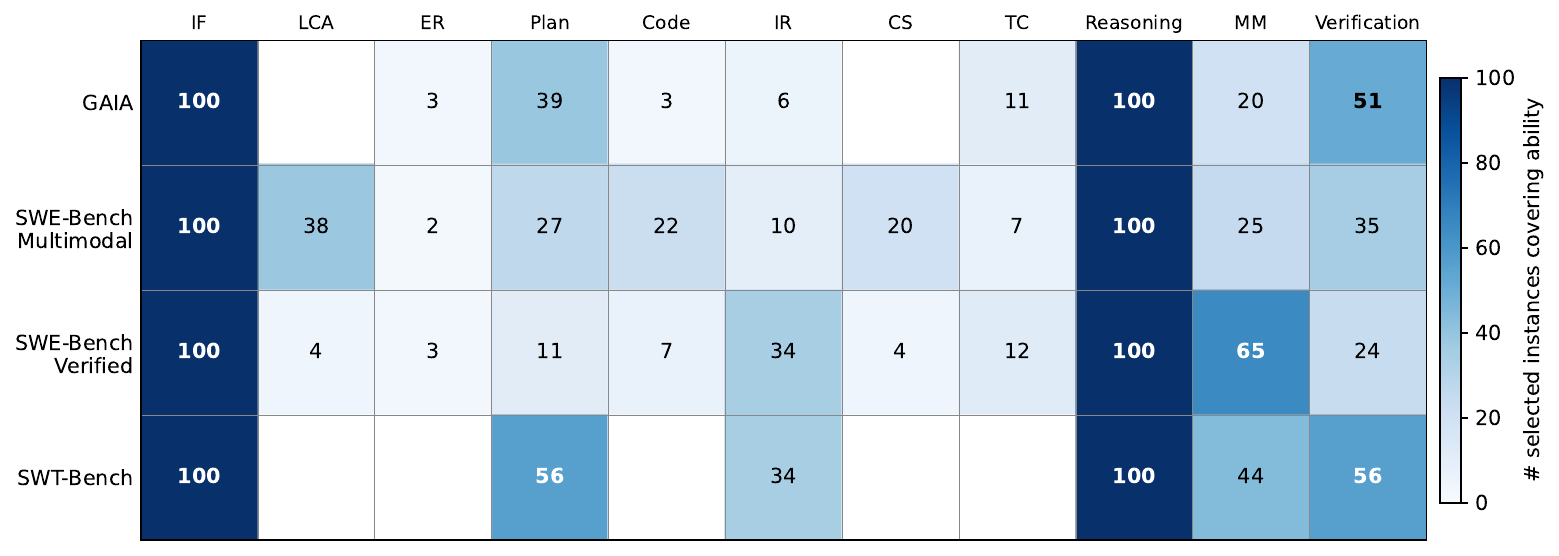}
    \label{fig:ability_heatmap}
    \vspace{-15pt}
\end{figure}

\autoref{fig:ability_heatmap} shows the ability distribution (ability categorization shown in \autoref{tab:benchmarks_combined}) of the $C=100$ selected source instances of \benchmark{}, revealing what each agentic target requires from its proxy.
The distribution for each source benchmark could be found in \autoref{app:allocation_heatmap}.

Instruction Following and Reasoning are saturated at 100 selected instances for all four targets, since every source benchmark in our pool covers these abilities. The remaining capabilities vary substantially across targets, showing what each agentic benchmark uniquely requires.

\textbf{GAIA: Instruction Following and Verification and Test.}
GAIA's allocation is overwhelmingly drawn from IF-focused benchmarks IFEval and PlanBench, along with a focus on Verification and Test. This matches GAIA's nature as browser-based question answering: each task specifies multi-clause answer-format and browsing constraints, so success hinges on rigorously following the prompt and self-checking that the produced answer satisfies every clause.

\textbf{SWE-Bench Verified: Multimodal Understanding, Information Retrieval, and Verification and Test.}
SWE-Bench Verified's proxy is dominated by Multimodal Understanding ($65$) and Information Retrieval ($34$), with Verification and Test ($24$) next. Although the target task itself is text-only patch generation, the selection leans on discriminative visual and retrieval instances (drawn from VisualPuzzles and VisualWebBench) that most sharply separate frontier models, together with a test-checking signal that mirrors running the hidden test suite; the direct code-editing abilities (Planning, Code Generation) contribute less because most calibration models already handle them and so carry little discriminative signal.

\textbf{SWE-Bench Multimodal: long-context aggregation.}
SWE-Bench Multimodal is led by Long-Context Aggregation ($38$), with Verification and Test ($35$), Planning ($27$), Multimodal Understanding ($25$), and Code Generation ($22$) close behind, reflecting tasks that integrate information across long issues, repository code, and screenshots. Notably, the Multimodal allocation ($25$) is moderate rather than dominant, because many SWE-bench Multimodal instances can be solved primarily from the textual issue and code with the visual contents merely as a supporting role.

\textbf{SWT-Bench: Verification and Test and Planning.}
SWT-Bench peaks on Verification and Test ($56$) and Planning ($56$), with sizable Multimodal Understanding ($44$) and Information Retrieval ($34$) contributions as well. The benchmark asks the model to author tests that exercise specific bug-triggering paths, which requires planning a multi-step test sequence and reasoning carefully about what each assertion checks, precisely the Verification and Test and Planning abilities.

Overall, \autoref{fig:ability_heatmap} suggests that the selection process could capture both a shared capability requirement of agentic tasks and benchmark-specific capability signatures.

\subsection{Bootstrap}
\label{sec:bootstrap_ablation}

To isolate the effect of the bootstrap pooling described in \autoref{sec:regression}, we perform an ablation by removing the bootstrap step, holding selection and regression fixed. \autoref{tab:no_bootstrap_results} reports the resulting LOOCV scores alongside their with-bootstrap counterparts from \autoref{tab:main_results}.

\begin{table}[htbp]
\centering
\small
\caption{No-bootstrap LOOCV results compared against the with-bootstrap LOOCV results in \autoref{tab:main_results}.
$\Delta$ MAE and $\Delta$ Spearman are computed as with-bootstrap minus no-bootstrap.}
\label{tab:no_bootstrap_results}
\begin{tabular}{l c c c c}
\toprule
\textbf{Target} & \textbf{No-bootstrap MAE} & \textbf{No-bootstrap Sp.} & \textbf{$\Delta$ MAE} & \textbf{$\Delta$ Sp.} \\
\midrule
GAIA                 & 6.45\% & 0.71 & $-0.68\%$ & $+0.08$ \\
SWE-bench Verified   & 2.75\% & 0.39 & $-0.66\%$ & $+0.28$ \\
SWE-bench Multimodal & 3.40\% & 0.69 & $-1.17\%$ & $+0.20$ \\
SWT-bench            & 5.68\% & 0.82 & $-0.56\%$ & $+0.07$ \\
\midrule
\textbf{Average}     & \textbf{4.57\%} & \textbf{0.66} & $\mathbf{-0.77\%}$ & $\mathbf{+0.15}$ \\
\bottomrule
\end{tabular}
\end{table}

Bootstrap helps on every target: average MAE improves by $0.77\%$ and average Spearman by $+0.15$, with no target regressing on either metric. These results justify keeping the bootstrap as the default configuration of \system{}.


\section{Conclusion}
\label{sec:conclusion}

We presented \system{}, a framework for predicting agentic-benchmark performance from a small, automatically selected set of non-agentic source instances. \system{} combines noise-aware bootstrap regression with two complementary filter-based selection signals: SVD leverage as a target-independent geometric prior, and rank correlation as a target-specific relevance score. These signals are ensembled with a learned per-target weight, allowing \system{} to identify compact, informative proxy subsets for each agentic benchmark.

Across four agentic targets and \nummodels{} models, \system{} achieves $3.80\%$ MAE, $0.81$ Spearman correlation, and around $85\%$ pairwise accuracy under leave-one-out cross-validation, while costing roughly $100\times$ less than a random target-sampling baseline of matched quality. Beyond prediction accuracy, the selected source instances provide interpretable per-target capability profiles, revealing each agentic benchmark's distinctive ability mix without requiring human-supplied annotations.

These results suggest that \system{} can make agentic evaluation substantially more practical during model development. Developers can use it to cheaply rank candidate models before running full agentic evaluations, monitor training checkpoints or hyperparameter runs with much denser feedback, and gate expensive full-harness evaluations by first screening whether a model is likely to be competitive. In this way, \system{} reduces agentic evaluation from hours of harness setup and hundreds of API calls per model to minutes of static-benchmark evaluation at roughly $\frac{1}{100}$ of the cost.

The main limitation is that these use cases depend on the calibration models being representative of future models. When a new model falls outside the calibration distribution, or reflects a substantially different architecture or training paradigm, proxy error may increase. In practice, this means the calibration set should be refreshed periodically as model distributions shift. We hope \system{} makes agentic evaluation routinely affordable for model development and serves as a foundation for future work on automatic, scalable benchmarking of agentic capabilities.

\bibliography{reference}
\bibliographystyle{colm2026_conference}


\appendix
\section{Related Work}
\label{sec:related_work}
Although agentic benchmarks provide more direct measurements of agent behaviors, their multi-turn and execution-based nature introduces major practical bottlenecks. To reduce the significant cost of evaluation, a growing line of work studies how to approximate benchmark scores using fewer examples. 
Methods such as tinyBenchmarks~\citep{10.5555/3692070.3693466}, PSN-IRT~\citep{zhou2026lostbenchmarksrethinkinglarge}, metabench~\citep{kipnis2025metabench}, and SparseEval~\citep{zhang2026sparseeval} identify highly discriminative subsets from within a target benchmark, using item response theory, learned task representations, sparse optimization~\citep{perlitz-etal-2024-efficient}. 

Other approaches such as APTBench~\citep{qin2025aptbenchbenchmarkingagenticpotential} convert agentic trajectories into static multiple-choice questions, and SWE-bench Verified~\citep{chowdhury2024swebenchverified} reduces the original benchmark to a human-validated subset of 500 instances for more reliable evaluation. These methods aim to preserve signal about the original benchmark while lowering cost. All of these methods operate within a single benchmark distribution, while the problem of reliably predicting performance on an agentic benchmark from an external pool of heterogeneous and inexpensive evaluation benchmarks remains largely unaddressed~\citep{zhang2025how}.
\section{Benchmark and Capabilities}
\label{app:benchmark_details}

\autoref{tab:benchmarks_combined} provides a representative list of benchmarks from two settings: standard LLM-based static benchmarks ACPBench \citep{kokel2025acpbench}, AIME \citep{aime_2025}, BEIR \citep{thakur2021beir}, BFCL \citep{patil2025the}, DebugBench \citep{tian2024debugbench}, GPQA \citep{rein2024gpqa}, HumanEval \citep{chen2021evaluating}, IFEval \citep{zhou2023instructionfollowingevaluationlargelanguage}, InFoBench \citep{qin2024infobench}, LIFBench \citep{wu2025lifbench}, LiveCodeBench \citep{jain2025livecodebench}, LogiQA \citep{liu2020logiqa}, MBPP \citep{austin2021program}, MMLU \citep{hendrycks2020measuring}, MMMU \citep{yue2024mmmu}, PlanBench \citep{valmeekam2023planbench}, RepoBench \citep{liu2023repobench}, VisualPuzzles \citep{song2025visualpuzzles}, and VisualWebBench \citep{liu2024visualwebbench}; and agentic benchmarks including GAIA~\citep{mialon2024gaia}, SWE-Bench Verified~\citep{jimenez2024swebench}, SWE-Bench Multimodal~\citep{yang2025swebench}, SWT-Bench~\citep{mundler2024swtbench}. 

We organize model capabilities into \numabilities{} categories: instruction following (adhering to explicit constraints), long context aggregation (aggregation over extended inputs), error recovery (correcting intermediate mistakes), planning (sequencing multi-step actions), code generation (writing executable programs), information retrieval (locating relevant content), code search (navigating and understanding codebases), tool calling (invoking structured APIs), reasoning (logical and mathematical inference), multimodal understanding (interpreting visual inputs), verification and test (checking correctness of outputs and reasoning about test cases).

\section{Per-source-benchmark allocation}
\label{app:allocation_heatmap}

\begin{figure}[htbp]
    \centering
    \includegraphics[width=\textwidth]{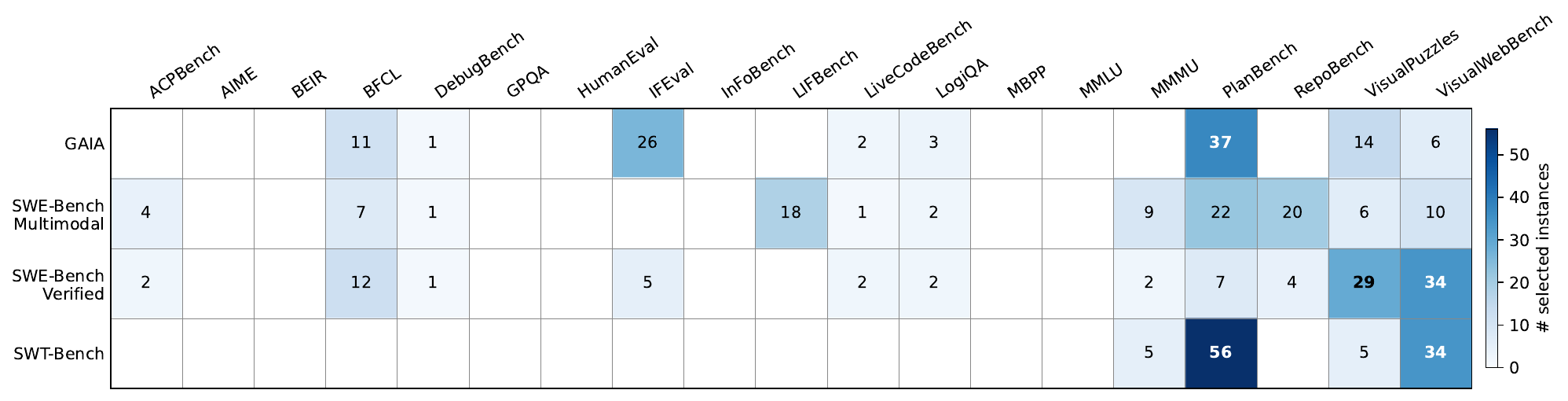}
    \caption{Number of source instances selected per (target, source benchmark) pair at $C=100$. Cells are deduplicated across the Local and Global selection lists, so each row sums to exactly $100$.}
    \label{fig:allocation_heatmap}
\end{figure}

\autoref{fig:allocation_heatmap} shows the same $C=100$ selected instances analyzed in \autoref{sec:allocation_analysis}, but disaggregated by the originating source benchmark rather than aggregated over the abilities each benchmark covers. This view exposes which specific source benchmarks \system{} draws from for each target, and complements the capability-level discussion in the main text.

\paragraph{Shared predictors across targets.}
A small set of source benchmarks is consistently selected across all four targets. PlanBench is selected heavily for every target except SWE-bench Verified ($37$ for GAIA, $7$ for SWE-bench Verified, $22$ for SWE-bench Multimodal, $56$ for SWT-bench), reflecting that planning--verification chains generalize across most agentic surfaces. VisualPuzzles ($14$/$29$/$6$/$5$) and VisualWebBench ($6$/$34$/$10$/$34$) are selected for every target, supplying multimodal-reasoning and web-interaction signals that discriminate frontier models even when the target task itself is unimodal.

\paragraph{Target-specific benchmark allocation.}
Beyond the shared substrate, each target draws on a distinctive set of source benchmarks that match its surface form:
\begin{itemize}[leftmargin=*]
    \item \textbf{SWE-bench Verified} concentrates on VisualWebBench ($34$) and VisualPuzzles ($29$), leaning on visual and web-interaction reasoning that sharply separates frontier models even though the target task is text-only patch generation.
    \item \textbf{SWE-bench Multimodal} draws on PlanBench ($22$), RepoBench ($20$), and InFoBench ($18$), reflecting its long-context navigation across code, screenshots, and issues.
    \item \textbf{GAIA} is dominated by PlanBench ($37$) and IFEval ($26$), matching its browser-based question-answering style that combines strict instruction following with multi-step planning.
    \item \textbf{SWT-bench} concentrates on PlanBench ($56$) and VisualWebBench ($34$), consistent with its emphasis on planning multi-step test generation and validating outputs against specifications.
\end{itemize}

\section{Budget Sweep}
\label{app:budget_sweep}

\begin{table}[htbp]
\centering
\caption{Goal A / Goal B LOOCV performance as a function of the source-instance budget $C$, averaged over the \numagentic{} agentic targets. Best value per column in bold.}
\label{tab:budget_sweep}
\begin{tabular}{c c c c}
\toprule
\textbf{$C$} & \textbf{Goal A MAE} & \textbf{Goal A Sp.} & \textbf{Goal B Acc.} \\
\midrule
25  & 4.02\% & 0.832 & 83.98\% \\
50  & 4.06\% & 0.794 & 83.00\% \\
100 & 3.80\% & 0.807 & 84.37\% \\
200 & 3.63\% & 0.832 & 86.76\% \\
300 & 3.51\% & 0.838 & 86.76\% \\
400 & \textbf{3.30\%} & \textbf{0.868} & 86.20\% \\
500 & 3.44\% & 0.855 & \textbf{89.27\%} \\
\bottomrule
\end{tabular}
\end{table}

\autoref{tab:budget_sweep} reports LOOCV performance as the source-instance budget $C$ is swept from $25$ to $500$, averaged over the \numagentic{} agentic targets. Goal A scales gracefully: MAE broadly improves as the budget grows, reaching its minimum at $C=400$ ($3.30\%$, with Spearman $0.868$) before bending back at $C=500$ ($3.44\%$), suggesting that the small calibration set starts to overfit when too many regressors are admitted. Goal B accuracy, by contrast, continues climbing all the way to $C=500$ ($89.27\%$), reflecting that pairwise ranking benefits from additional discriminating instances longer than absolute prediction does.

Even small budgets are highly informative: at $C=25$ \system{} already achieves $4.02\%$ MAE, $0.832$ Spearman, and $83.98\%$ pair accuracy---within $\sim 0.7\%$ MAE of the $C=400$ optimum and within $\sim 5\%$ pair accuracy of the $C=500$ optimum, while running at a fraction of the eval cost (\autoref{fig:cost_curve}). $C=100$, the headline budget used elsewhere in the paper, sits at a practical sweet spot: it remains competitive with the larger budgets while keeping per-model evaluation cost low.

\section{Lasso and Ridge Baseline}
\label{app:sparse_baseline}

We compare \system{} to two baselines that perform selection and prediction simultaneously: (1) an $\ell_1$-penalized \textbf{Lasso} \citep{tibshirani1996lasso} whose non-zero columns form the selected subset, and (2) an $\ell_2$-penalized \textbf{Ridge} \citep{hoerl1970ridge} whose top-$C$ columns by $|w_i|$ form the subset (with Ridge then refit on those $C$ columns). Per LOOCV fold, each baseline is fit on the $|M_\mathrm{train}|$ training models; the resulting selection is then reused as the input to a separate pair-logistic for Goal B (mirroring \system{}'s pinned-selection protocol). To give each baseline its fairest chance, we sweep the regularization strength $\alpha \in {10^{-5}, 10^{-4}, 10^{-3}, 10^{-2}, 10^{-1}}$ \emph{per target} and report each baseline at the $\alpha^\star$ that minimises its own LOOCV MAE.

\begin{table}[htbp]
\centering
\caption{Comparison of \system{} against two baselines, Lasso (L1) and Ridge (L2) with top-$C$ ($C=100$) selection by $|w_i|$ on Goal A (absolute score; MAE / Spearman / Pearson) and Goal B (pairwise accuracy). For each baseline, the regularization strength $\alpha^\star$ is selected per target from the grid $\alpha \in \{10^{-5}, 10^{-4}, 10^{-3}, 10^{-2}, 10^{-1}\}$ to minimise LOOCV MAE; the column $\alpha^\star$ reports the chosen value. Left: in-sample fit on all \nummodels{} models; right: strict LOOCV.}
\label{tab:sparse_baseline}
\resizebox{\textwidth}{!}{
\begin{tabular}{l l c cccc cccc}
\toprule
 &  &  & \multicolumn{4}{c}{\textbf{In-sample fit}} & \multicolumn{4}{c}{\textbf{LOOCV}} \\
\cmidrule(lr){4-7} \cmidrule(lr){8-11}
 &  &  & \multicolumn{3}{c}{\textbf{Goal A}} & \textbf{Goal B} & \multicolumn{3}{c}{\textbf{Goal A}} & \textbf{Goal B} \\
\cmidrule(lr){4-6} \cmidrule(lr){7-7} \cmidrule(lr){8-10} \cmidrule(lr){11-11}
\textbf{Target} & \textbf{Method} & $\alpha^\star$ & \textbf{MAE} & \textbf{Sp.} & \textbf{Pear.} & \textbf{Acc.} & \textbf{MAE} & \textbf{Sp.} & \textbf{Pear.} & \textbf{Acc.} \\
\midrule
\multirow{3}{*}{GAIA}                 & \system{} & ---      & 4.58\% & 0.85 & 0.90 & 92.78\% & 5.77\% & 0.79 & 0.78 & 83.33\% \\
                                      & Lasso     & $10^{-4}$ & 0.02\% & 1.00 & 1.00 & 99.45\% & 8.62\% & 0.38 & 0.40 & 81.87\% \\
                                      & Ridge     & $10^{-1}$ & 1.07\% & 0.98 & 0.99 & 95.60\% & 7.10\% & 0.79 & 0.75 & 86.26\% \\
\addlinespace[2pt]
\multirow{3}{*}{SWE-bench Verified}   & \system{} & ---      & 1.34\% & 0.94 & 0.87 & 93.26\% & 2.09\% & 0.67 & 0.61 & 78.54\% \\
                                      & Lasso     & $10^{-5}$ & 0.00\% & 1.00 & 1.00 & 98.90\% & 2.82\% & 0.15 & 0.16 & 71.98\% \\
                                      & Ridge     & $10^{-1}$ & 0.19\% & 1.00 & 1.00 & 97.80\% & 2.71\% & 0.66 & 0.39 & 70.88\% \\
\addlinespace[2pt]
\multirow{3}{*}{SWE-bench Multimodal} & \system{} & ---      & 1.47\% & 0.94 & 0.92 & 93.26\% & 2.23\% & 0.89 & 0.80 & 85.39\% \\
                                      & Lasso     & $10^{-2}$ & 2.00\% & 0.97 & 0.92 & 91.21\% & 3.53\% & 0.60 & 0.57 & 79.67\% \\
                                      & Ridge     & $10^{-4}$ & 1.00\% & 0.95 & 0.94 & 92.86\% & 2.79\% & 0.80 & 0.83 & 82.42\% \\
\addlinespace[2pt]
\multirow{3}{*}{SWT-bench}            & \system{} & ---      & 4.14\% & 0.92 & 0.89 & 93.96\% & 5.12\% & 0.89 & 0.78 & 90.11\% \\
                                      & Lasso     & $10^{-5}$ & 0.00\% & 1.00 & 1.00 & 100.00\% & 7.77\% & 0.70 & 0.61 & 84.62\% \\
                                      & Ridge     & $10^{-1}$ & 1.04\% & 0.94 & 0.99 & 93.96\% & 8.03\% & 0.78 & 0.64 & 82.97\% \\
\midrule
\multirow{3}{*}{\textbf{Average}}     & \system{} & ---      & \textbf{2.88\%} & \textbf{0.91} & \textbf{0.90} & \textbf{93.31\%} & \textbf{3.80\%} & \textbf{0.81} & \textbf{0.74} & \textbf{84.37\%} \\
                                      & Lasso     & ---      & \textbf{0.51\%} & \textbf{0.99} & \textbf{0.98} & \textbf{97.39\%} & \textbf{5.69\%} & \textbf{0.46} & \textbf{0.43} & \textbf{79.53\%} \\
                                      & Ridge     & ---      & \textbf{0.82\%} & \textbf{0.97} & \textbf{0.98} & \textbf{95.05\%} & \textbf{5.16\%} & \textbf{0.76} & \textbf{0.65} & \textbf{80.63\%} \\
\bottomrule
\end{tabular}
}
\end{table}

\autoref{tab:sparse_baseline} compares \system{} and the two baselines under identical in-sample fit and LOOCV across the agentic targets. We highlight three findings.

\paragraph{Both baselines overfit substantially more than PACE.}
Even with $\alpha$ tuned per target, both baselines fit near-perfectly in-sample but degrade sharply under LOOCV. Lasso achieves only $0.51\%$ in-sample MAE and $0.99$ Spearman on average, but drops to $5.69\%$ LOOCV MAE and $0.46$ Spearman. Ridge shows a similar pattern, increasing from $0.82\%$ to $5.16\%$ MAE and decreasing from $0.97$ to $0.76$ Spearman. By contrast, \system{} has a much smaller in-sample-to-LOOCV degradation: its MAE increases from $2.88\%$ to $3.80\%$, while Spearman decreases from $0.91$ to $0.81$. The pair preference accuracies tell the same story: \system{} drops from $93.31\%$ to $84.37\%$, whereas Lasso and Ridge fall from $97.39\%$ to $79.53\%$ and from $95.05\%$ to $80.63\%$, respectively. These gaps indicate that Lasso and Ridge can exploit the small set of models to obtain overly optimistic in-sample fits, while \system{}'s decoupled filter scoring yields selections that generalize better across held-out models.

\paragraph{PACE gives the best LOOCV prediction despite weaker in-sample fitting.}
The baselines' near-perfect in-sample scores do not translate into better held-out performance. Although \system{} is less aggressively fitted in-sample, it achieves the lowest average LOOCV MAE ($3.80\%$ vs.\ $5.69\%$ for Lasso and $5.16\%$ for Ridge), the highest LOOCV Spearman ($0.81$ vs.\ $0.46$ and $0.76$), and the highest LOOCV pairwise accuracy ($84.37\%$ vs.\ $79.53\%$ and $80.63\%$). This suggests that the strongest in-sample fit is not the right objective for model-efficient benchmark prediction; instead, robust source selection is critical.

\paragraph{The conclusion is robust to $\alpha$.}
\autoref{tab:sparse_sweep} reports the average LOOCV MAE and Spearman of each baseline at five different values of $\alpha$. \system{} outperforms both baselines at every $\alpha$ tested.

\begin{table}[htbp]
\centering
\small
\caption{Average LOOCV performance (across the \numagentic{} agentic targets) at each $\alpha$ on the sweep grid. PACE outperforms both baselines at \emph{every} $\alpha$ tested, with the best-$\alpha$ row reproduced from \autoref{tab:sparse_baseline} for reference.}
\label{tab:sparse_sweep}
\begin{tabular}{c cc cc}
\toprule
 & \multicolumn{2}{c}{\textbf{Lasso}} & \multicolumn{2}{c}{\textbf{Ridge (top-$C$)}} \\
\cmidrule(lr){2-3} \cmidrule(lr){4-5}
$\alpha$ & MAE & Sp. & MAE & Sp. \\
\midrule
$10^{-5}$ & 5.83\% & 0.44 & 8.83\% & 0.72 \\
$10^{-4}$ & 5.84\% & 0.41 & 7.30\% & 0.72 \\
$10^{-3}$ & 5.85\% & 0.40 & 6.11\% & 0.73 \\
$10^{-2}$ & 6.63\% & 0.28 & 5.88\% & 0.71 \\
$10^{-1}$ & 7.49\% & 0.00  & 5.18\% & 0.75 \\
\midrule
best $\alpha^\star$ (per target) & \textbf{5.69\%} & \textbf{0.46} & \textbf{5.16\%} & \textbf{0.76} \\
\bottomrule
\end{tabular}
\end{table}

\section{Limitations}

\textbf{Proxy gaming.}
Because \system{} selects a fixed, public set of proxy instances, a model developer who knows the proxy set could optimize specifically for those instances-achieving high proxy scores without improving genuine agentic capability.
This risk is analogous to benchmark contamination and grows as the proxy set becomes widely known.
Mitigations include periodically refreshing the proxy set, keeping it private until evaluation, or sampling a fresh subset at evaluation time.

\textbf{Small calibration set.}
Our experiments use \nummodels{} frontier models as the calibration set, which is small relative to the $C=100$ feature dimensionality.
Ridge regularization partially addresses this, but the learned regression weights may not be reliable for individual instances; they are better interpreted as aggregate signals.
As more models are evaluated on the source benchmarks, the calibration set will grow and selection quality is expected to improve.

\textbf{Coverage of agentic benchmarks.}
We evaluate on four agentic benchmarks sharing a common agent framework (OpenHands).
Whether \system{} generalizes to agentic benchmarks with different scaffolds, tool sets, or evaluation protocols (e.g., browser-based or embodied agents) remains to be studied.

\textbf{Static source pool.}
The \numsources{} source benchmarks were chosen to cover a broad range of capabilities, but the proxy set is only as good as the source pool.
If a target agentic benchmark requires capabilities not measured by any source benchmark, \system{} cannot recover the missing signal regardless of which instances it selects.


\newpage
\section*{NeurIPS Paper Checklist}

\begin{enumerate}

\item {\bf Claims}
    \item[] Question: Do the main claims made in the abstract and introduction accurately reflect the paper's contributions and scope?
    \item[] Answer: \answerYes{}
    \item[] Justification: The abstract and introduction claim that \system{} predicts agentic-benchmark performance from a small set of non-agentic instances at substantially lower cost; these claims are supported by the LOOCV results in \autoref{sec:main_results} (\autoref{tab:main_results}) and the cost--quality comparison in \autoref{fig:cost_curve}.

\item {\bf Limitations}
    \item[] Question: Does the paper discuss the limitations of the work performed by the authors?
    \item[] Answer: \answerYes{}
    \item[] Justification: \autoref{sec:discussion} contains a Limitations paragraph explicitly noting that predictions rely on the calibration models being representative of future models and that proxy error may grow under distribution shift.

\item {\bf Theory assumptions and proofs}
    \item[] Question: For each theoretical result, does the paper provide the full set of assumptions and a complete (and correct) proof?
    \item[] Answer: \answerNA{}
    \item[] Justification: The paper does not include formal theorems; the regression and selection procedures are described as concrete algorithms in \autoref{sec:method}.

    \item {\bf Experimental result reproducibility}
    \item[] Question: Does the paper fully disclose all the information needed to reproduce the main experimental results of the paper to the extent that it affects the main claims and/or conclusions of the paper (regardless of whether the code and data are provided or not)?
    \item[] Answer: \answerYes{}
    \item[] Justification: \autoref{sec:method} fully specifies the regression and selection algorithms, and \autoref{sec:experiments} reports the protocol (LOOCV over \nummodels{} models, $C=100$, bootstrap $B=300$, fixed seed, \numsources{} source benchmarks evaluated via lm-evaluation-harness and \numagentic{} agentic targets evaluated via the OpenHands SDK).

\item {\bf Open access to data and code}
    \item[] Question: Does the paper provide open access to the data and code, with sufficient instructions to faithfully reproduce the main experimental results, as described in supplemental material?
    \item[] Answer: \answerYes{}
    \item[] Justification: We will release an anonymized code repository containing the selection, regression, and evaluation pipelines, along with the per-model source-benchmark and target-benchmark score matrices, accompanied by exact commands to reproduce \autoref{tab:main_results} and \autoref{fig:cost_curve}.

\item {\bf Experimental setting/details}
    \item[] Question: Does the paper specify all the training and test details (e.g., data splits, hyperparameters, how they were chosen, type of optimizer) necessary to understand the results?
    \item[] Answer: \answerYes{}
    \item[] Justification: \autoref{sec:experiments} specifies the LOOCV split, the budget $C$ and bootstrap count $B$, the per-target ensemble weight tuning, and the regularization protocol; the methodology in \autoref{sec:method} defines the linear least-squares (Goal~A) and logistic (Goal~B) regressors and how their hyperparameters are auto-tuned via held-out evaluation.

\item {\bf Experiment statistical significance}
    \item[] Question: Does the paper report error bars suitably and correctly defined or other appropriate information about the statistical significance of the experiments?
    \item[] Answer: \answerNo{}
    \item[] Justification: The headline LOOCV numbers in \autoref{tab:main_results} are point estimates without error bars; we do, however, account for target-instance label noise via the bootstrap pooling described in \autoref{sec:regression} and report a paired with-vs-without-bootstrap ablation in \autoref{tab:no_bootstrap_results}, which addresses the dominant source of variance in our regime.

\item {\bf Experiments compute resources}
    \item[] Question: For each experiment, does the paper provide sufficient information on the computer resources (type of compute workers, memory, time of execution) needed to reproduce the experiments?
    \item[] Answer: \answerYes{}
    \item[] Justification: The \system{} regression and selection pipeline is light-weight (CPU-only; an SVD on a $14\times 44{,}238$ matrix and per-target ridge fits) and runs in minutes on a commodity laptop; the upstream source-benchmark scoring is done with lm-evaluation-harness and the agentic targets with the OpenHands SDK, with per-model dollar costs reported in \autoref{fig:cost_curve}.

\item {\bf Code of ethics}
    \item[] Question: Does the research conducted in the paper conform, in every respect, with the NeurIPS Code of Ethics \url{https://neurips.cc/public/EthicsGuidelines}?
    \item[] Answer: \answerYes{}
    \item[] Justification: The work uses publicly available benchmarks and frontier-model APIs in their intended evaluation regime; it does not involve human subjects, sensitive data, or release of high-risk artifacts.

\item {\bf Broader impacts}
    \item[] Question: Does the paper discuss both potential positive societal impacts and negative societal impacts of the work performed?
    \item[] Answer: \answerYes{}
    \item[] Justification: \autoref{sec:discussion} discusses positive impacts---reducing the cost of agentic evaluation makes rigorous benchmarking more accessible and supports denser model-development monitoring---and notes that proxy predictions can drift under distribution shift, which could mislead deployment decisions if the proxy is over-trusted in place of full evaluation.

\item {\bf Safeguards}
    \item[] Question: Does the paper describe safeguards that have been put in place for responsible release of data or models that have a high risk for misuse (e.g., pre-trained language models, image generators, or scraped datasets)?
    \item[] Answer: \answerNA{}
    \item[] Justification: The released artifacts are evaluation utilities and aggregate score matrices over public benchmarks; we do not release any pretrained models, generative systems, or scraped datasets that pose misuse risk.

\item {\bf Licenses for existing assets}
    \item[] Question: Are the creators or original owners of assets (e.g., code, data, models), used in the paper, properly credited and are the license and terms of use explicitly mentioned and properly respected?
    \item[] Answer: \answerYes{}
    \item[] Justification: All \numsources{} source benchmarks and \numagentic{} agentic targets are cited at first mention in \autoref{tab:benchmarks_combined} and used under their respective public licenses; the lm-evaluation-harness and OpenHands SDK are likewise cited as the evaluation infrastructure.

\item {\bf New assets}
    \item[] Question: Are new assets introduced in the paper well documented and is the documentation provided alongside the assets?
    \item[] Answer: \answerYes{}
    \item[] Justification: The released code repository will include documentation for the \system{} pipeline (selection, regression, evaluation), the per-model source/target score matrices, and example commands reproducing each table and figure in the paper.

\item {\bf Crowdsourcing and research with human subjects}
    \item[] Question: For crowdsourcing experiments and research with human subjects, does the paper include the full text of instructions given to participants and screenshots, if applicable, as well as details about compensation (if any)?
    \item[] Answer: \answerNA{}
    \item[] Justification: The paper does not involve crowdsourcing or human subjects; all data are model evaluations on existing public benchmarks.

\item {\bf Institutional review board (IRB) approvals or equivalent for research with human subjects}
    \item[] Question: Does the paper describe potential risks incurred by study participants, whether such risks were disclosed to the subjects, and whether Institutional Review Board (IRB) approvals (or an equivalent approval/review based on the requirements of your country or institution) were obtained?
    \item[] Answer: \answerNA{}
    \item[] Justification: The paper does not involve human subjects, so IRB review is not applicable.

\item {\bf Declaration of LLM usage}
    \item[] Question: Does the paper describe the usage of LLMs if it is an important, original, or non-standard component of the core methods in this research? Note that if the LLM is used only for writing, editing, or formatting purposes and does \emph{not} impact the core methodology, scientific rigor, or originality of the research, declaration is not required.
    \item[] Answer: \answerNA{}
    \item[] Justification: LLMs are the \emph{subject} of evaluation in this work, not a component of the proposed method. The core \system{} algorithm (SVD-based filter selection plus linear/logistic regression) does not involve LLMs in any non-standard way.

\end{enumerate}

\end{document}